\documentclass[letterpaper]{article} 
\usepackage{aaai24}  
\usepackage{times}  
\usepackage{helvet}  
\usepackage{courier}  
\usepackage[hyphens]{url}  
\usepackage{graphicx} 
\usepackage{natbib}  
\usepackage{caption} 
\usepackage[T1]{fontenc}

\setcitestyle{authoryear}

\usepackage[utf8]{inputenc} 
\usepackage{amsfonts}       
\usepackage{nicefrac}       
\usepackage{microtype}      
\usepackage{xcolor}         
\definecolor{bluecite}{HTML}{0875b7}

\usepackage{subcaption}
\usepackage{comment}
\usepackage{amsmath,amssymb}
\usepackage{booktabs}
\usepackage{longtable}
\usepackage{fancyvrb}
\usepackage{algorithm}
\usepackage{algorithmic}
\usepackage[export]{adjustbox}

\usepackage{newfloat}
\usepackage{listings}
\DeclareCaptionStyle{ruled}{labelfont=normalfont,labelsep=colon,strut=off} 
\floatstyle{ruled}
\newfloat{listing}{tb}{lst}{}
\floatname{listing}{Listing}
\title{On Diagnostics for Understanding\\ Agent Training Behaviour in Cooperative MARL}
\author {
    Wiem Khlifi\textsuperscript{\rm 1, \rm 2},
    Siddarth Singh\textsuperscript{\rm 1},
    Omayma Mahjoub\textsuperscript{\rm 1},
    Ruan de Kock\textsuperscript{\rm 1}
    Abidine Vall\textsuperscript{\rm 3}
    Rihab Gorsane\textsuperscript{\rm 1}
    Arnu Pretorius\textsuperscript{\rm 1}
}
\affiliations {
    \textsuperscript{\rm 1}InstaDeep Ltd\\
    \textsuperscript{\rm 2}National School of Computer Science, Tunisia\\
    \textsuperscript{\rm 3}National School of Engineering of Tunis\\
}

\begin{document}
\maketitle
\begin{abstract}

Cooperative multi-agent reinforcement learning (MARL) has made substantial strides in addressing the distributed decision-making challenges. However, as multi-agent systems grow in complexity, gaining a comprehensive understanding of their behaviour becomes increasingly challenging. Conventionally, tracking team rewards over time has served as a pragmatic measure to gauge the effectiveness of agents in learning optimal policies. Nevertheless, we argue that relying solely on the empirical returns may obscure crucial insights into agent behaviour. In this paper, we explore the application of explainable AI (XAI) tools to gain profound insights into agent behaviour. We employ these diagnostics tools within the context of Level-Based Foraging and Multi-Robot Warehouse environments and apply them to a diverse array of MARL algorithms. We demonstrate how our diagnostics can enhance the interpretability and explainability of MARL systems, providing a better understanding of agent behaviour.

\end{abstract}

\section{Introduction}
    
Multi-agent Reinforcement Learning (MARL) has shown immense promise in tackling complex decision-making challenges across various domains, such as robotics, healthcare, and energy. For example, MARL has been used to coordinate robotic surgery teams \citep{roboticsurgeryieee2021}, optimise smart grid operations \citep{Roesch2020SmartGF}, and manage air traffic flow \citep{IntelligentTrafficControl}.  

Standard methods for evaluating MARL in the cooperative setting rely solely on measuring empirical returns or statistics derived from them \citep{gorsane2022standardised}. Although tracking returns over time is a reasonable way to measure whether agents are learning optimal policies, recent scrutiny has revealed the limitations of evaluation methods, prompting the development of more robust solutions \citep{Whiteson2011ProtectingAE, agarwal2022deep, gorsane2022standardised}. Yet these standard measures often fail to explain the underlying behaviour of MARL algorithms \citep{canese2021multi, disentanglingson22a}. Addressing this challenge is where the growing field of Explainable Artificial Intelligence (XAI) steps in.


XAI consists of machine learning (ML) techniques that can provide human interpretable insights into the inner workings of ML models \citep{arrieta2020explainable}. It has been extensively explored in single-agent RL  \citep{heuillet2021explainability, hu2022policy, vouros2022explainable, puiutta2020explainable, dazeley2023explainable}, where various methods have been proposed to analyse and visualise the policies, value functions, and salient features of individual agents. However, explainability remains underrepresented in the context of MARL, where agents interact with each other and the environment in complex and dynamic ways. Existing tools for MARL explainability are limited and specialised for specific scenarios or algorithms \citep{shapley1953, heuillet2022collective}. Therefore, there is a need for more general and robust tools that can enhance interpretability and understanding of MARL systems.

In this paper, we argue that it is necessary to not rely on empirical performance for the evaluation of MARL algorithms; instead, we can supplement performance analysis with XAI tools to provide direct insights into agent behaviour to provide a more holistic view of results. To accomplish this, we adapt multiple simple implementation-agnostic methods for analyzing behaviour and we demonstrate how they can detect and offer valuable insights.
Specifically, we assess learning stability using Policy Entropy and Agent Update Divergence and analyse a Task Switching metric to provide insights into the learned action distributions of MARL agents. From this analysis in the LBF and RWARE domains, we uncover insights into how premature convergence limits the performance of MAPPO on LBF compared to MAA2C and show how the limitations of Q-learning methods on RWARE are not purely a byproduct of sparsity as originally hypothesised by \citep{papoudakis2021benchmarking}.

\section{Diagnostics Tools}
To assess agents' behaviour during training or evaluation, we investigate a diverse set of tools used as diagnostics of policy learning. We provide an overview of these tools which are typically based on metrics derived from prior work that we adapt to the MARL setting.

\textbf{Policy Entropy. }
\citep{abdallah2009global} demonstrated that only tracking global returns may mask potential policy instability issues in agents. To address this, the authors proposed using Policy Entropy as a measure of stability during learning. 
This metric is used to measure the uncertainty or randomness of a stochastic policy of an agent $i$, which is a probability distribution over actions given an observation and we can compute the Policy Entropy of agent $i$'s policy $\pi_i$ as follows
\begin{equation}
H(\pi_i) = - \sum_{a_i \in A_i} \pi_i(a_i|o_i) \log \pi_i(a_i|o_i),    
\end{equation}
where $o_i$ is the agent's observation and $a_i \in A_i$ denotes an action from the agent's action set $A_i$.\\
Intuitively, this measures how spread out the probabilities, given by a policy $\pi_i$, are across different actions. If the probabilities are evenly distributed (i.e., the policy is highly uncertain), the entropy is higher and it indicates that the agent's choices are more diverse and exploratory, as it is not favouring any specific action strongly. Contrarily, if the probabilities are concentrated on a few actions (i.e., the policy is more deterministic), the entropy is lower and it indicates that the agent tends to concentrate its probability on a smaller subset of actions. 


\textbf{Agent Update Divergence. }
The per-step KL divergence is used as a metric to assess the policy changes of each agent throughout the training process. It provides insights into the stochastic nature of the policy, serving as a monitor of the evolution of the policy.  
A persistent high KL divergence observed over time indicates a propensity for a stochastic policy that persists in exploring varied strategies. Conversely, a low KL divergence implies that the agent’s policy is trending towards predictability, converging either towards a deterministic behavioural or, possibly, a suboptimal strategy.
If we define the cross-entropy between two policies $\pi$ and $\pi^{\prime}$ as $H(\pi, \pi^{\prime}) = - \sum_{a \in A} \pi(a|o) \log \pi^{\prime}(a|o))$, the update divergence for each agent $i$ is given by
\begin{equation}
D_{kl}(\pi_i \| \pi^{\prime}_i) = H(\pi_i, \pi^{\prime}_{i}) - H(\pi_i), 
\end{equation}
where $\pi_i$ and $\pi^{\prime}_{i}$ are the current and old policy of agent $i$.

\textbf{Task Switching. }
During training agents often assume distinct roles \citep{wang2020roma,phan2021vast}. While some MARL methods such as ROMA \citep{wang2020roma} explicitly account for this, it is not a typical feature of algorithms. Although the above KL divergence has been proposed as a measure of agent action diversity, it can vary even among agents with similar policies due to observation differences \citep{hu2022policy}. For example, in a football game, defenders with identical roles may respond differently based on unique observations \citep{kurach2020google}.
Therefore, we explore implementation-agnostic methods to monitor agent diversity in MARL training by using a Task Switching tool. We implement such a tool by tracking the usage of each agent's action at each timestep $t$ during the evaluation phase and computing the overall likelihood of each agent's action being selected as
\begin{equation}
\mathcal{T}_i = \sigma\left( \sum^{T}_{t=0} \mathbf{a}^{t}_{i}\right),
\end{equation}
where $\mathbf{a}^{t}_{i}$ represents the one-hot encoded vector of the action chosen by agent $i$ at timestep $t$ with $T$ being the terminal timestep of the evaluation phase and $\sigma$ the softmax function.

A Task Switching metric makes it possible to quantify the frequency at which an agent selects specific actions during the evaluation phase and it can be useful in diagnosing the learning of agents by measuring how often they use the different actions. 

The Task Switching tool is complementary to the Policy Entropy and Agent Update Divergence tools, since it measures the actual behaviour of the agents after acquiring some knowledge, during the evaluation phase, rather than during the learning phase in addition to the fact that the actual behaviour of the agents depends not only on their policies, but also on their observations and roles in the environment.

\section{Experiment}

\begin{figure}[ht!]
    \centering
      \includegraphics[width=0.26\textwidth]{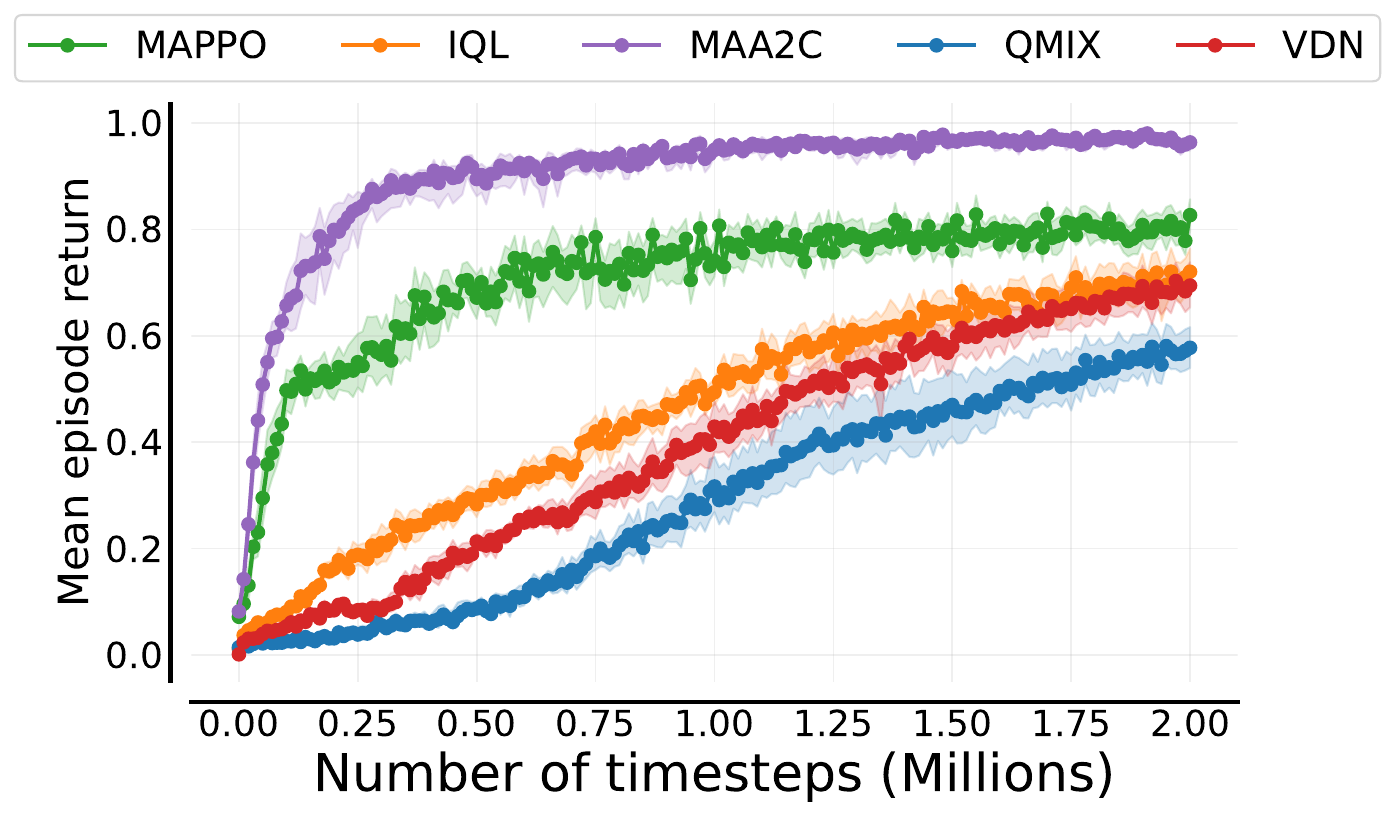}
    \includegraphics[width=0.2\textwidth]{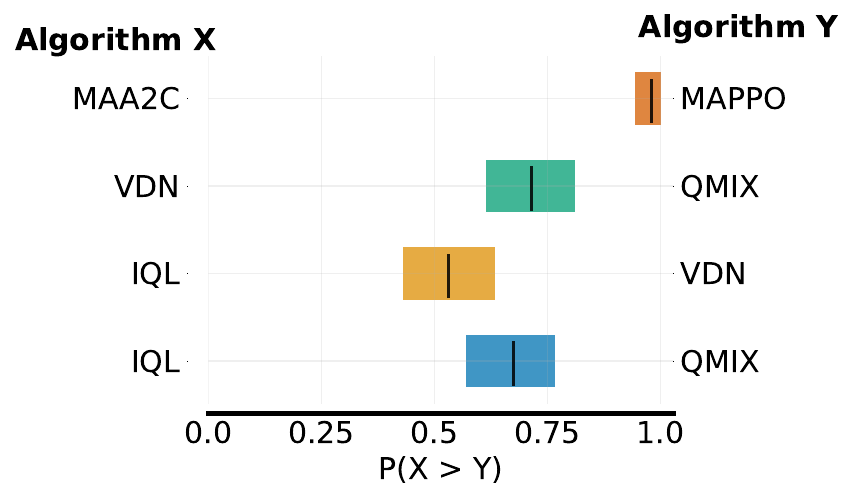}
    \includegraphics[width=0.26\textwidth]{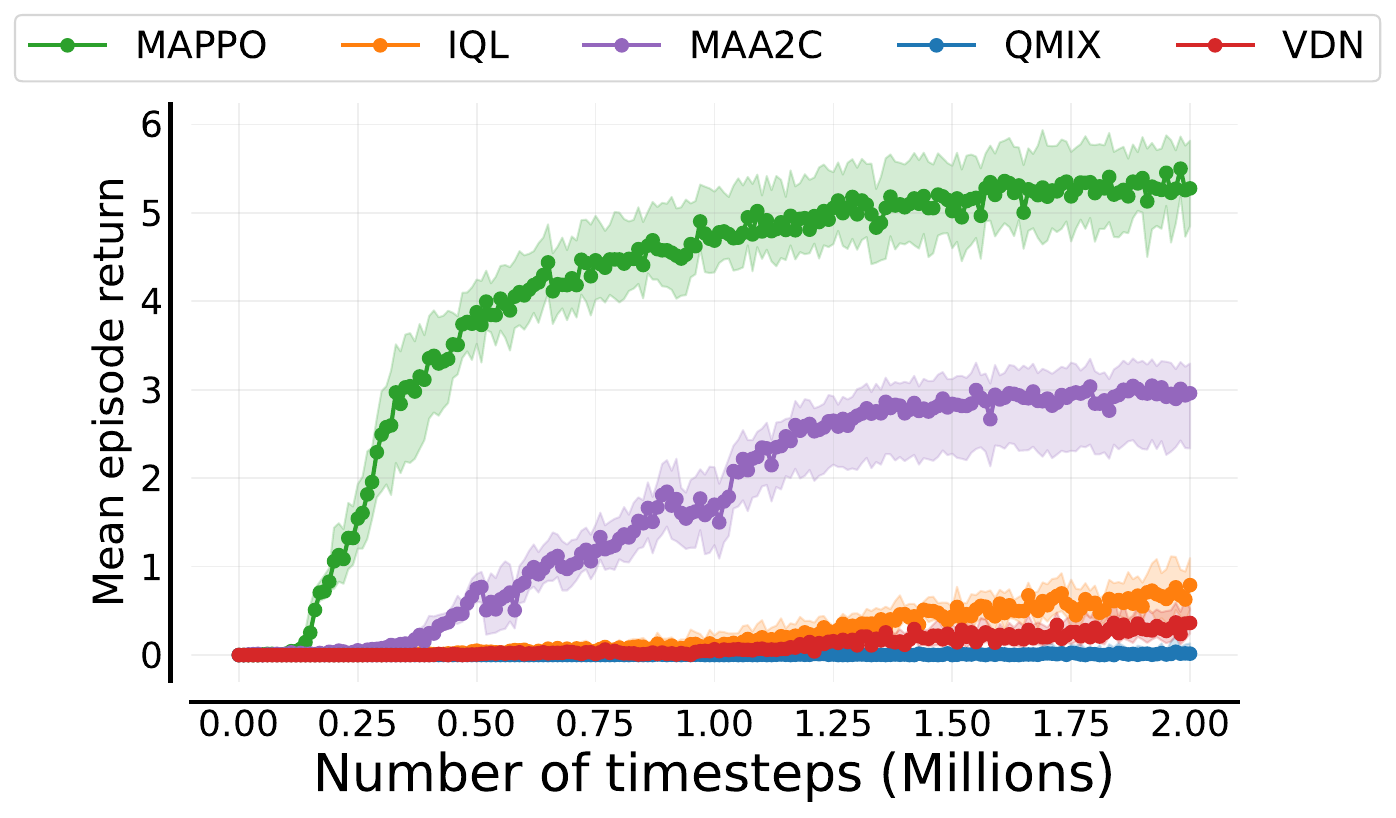}
    \includegraphics[width=0.2\textwidth]{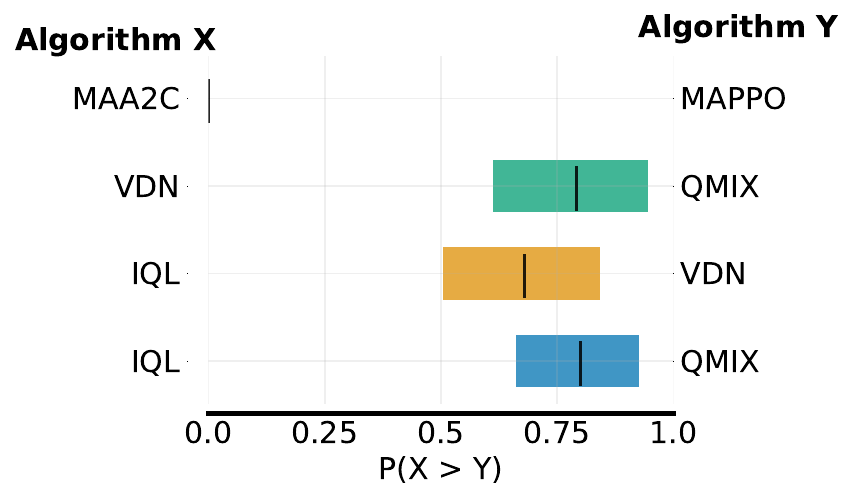}
    \caption{\textit{Algorithm performance on LBF and RWARE including probability of improvement(right), and sample efficiency curves(left)}. \textbf{Top row:} Performance of algorithms on 7 LBF tasks. \textbf{Bottom row:} Performance of all algorithms on 3 RWARE tasks.}
    \label{fig:benchmarking_paper_redo}
\end{figure}

We conducted experiments using the same training setup as \cite{papoudakis2021benchmarking}. We used the optimized hyperparameters and the EPyMARL framework, following the evaluation protocol proposed by \cite{gorsane2022standardised}. In addition, we trained a range of algorithms, from Q-learning to policy gradient (PG) methods, across seven distinct scenarios in the Level-Based Foraging (LBF) environment and three distinct scenarios in the Multi-Robot Warehouse (RWARE), using 10 different seeds. Experiment results are depicted in Figure \ref{fig:benchmarking_paper_redo}.

\textbf{Environment.} In our study, we directed our attention toward the two environments, LBF and RWARE,  as outlined in prior works \citep{albrecht2015gametheoretic,albrecht2019reasoning,papoudakis2021benchmarking}. In these environments, the reward signal is notably sparse, rendering it a more challenging task during the learning phase compared to environments like SMAC \citep{SMAC, papoudakis2021benchmarking}.

\begin{figure}
\centering
\includegraphics[width=0.55\linewidth]{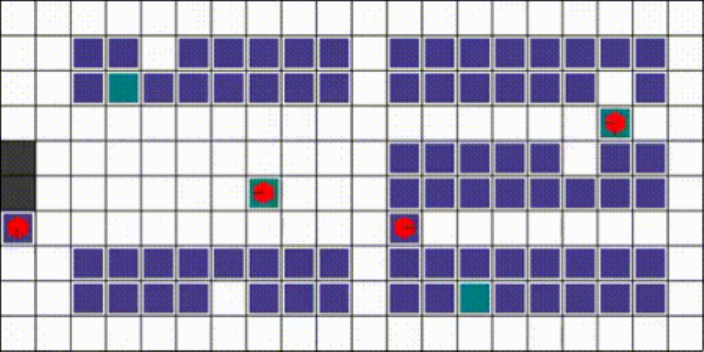} 
\includegraphics[width=0.35\linewidth]{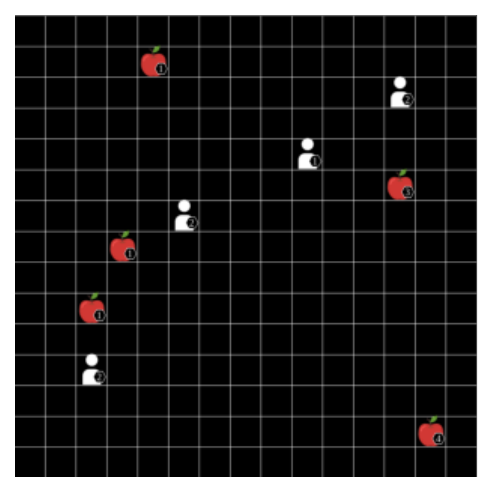}
\caption{\textbf{Top}: Multi-Robot Warehouse (RWARE). \textbf{Bottom}: Level-based foraging (LBF)}
\label{fig:rware_lbf}
\end{figure}

\textit{LBF} represents a mixed cooperative-competitive environment that emphasises coordination between agents. As illustrated in Figure \ref{fig:rware_lbf}, agents are placed within a grid world and assigned different levels. To collect food, the cumulative level of participating agents must meet or exceed the food's designated level. Agents receive points equivalent to the level of the collected food and their own level.\\
\textit{RWARE} \citep{christianos2020shared, papoudakis2021benchmarking} is a multi-agent environment that is designed to represent a simplified setting where robots move goods around a warehouse. This environment requires agents (circles) to move requested shelves (green squares) to the goal (dark squares) and back to an empty square as illustrated at the top of Figure \ref{fig:rware_lbf}. Tasks are partially observable with a very sparse reward signal as agents have limited sight and are rewarded only upon successful delivery.

\textbf{Algorithms.} Our analysis focuses on a subset of commonly used MARL algorithms, including both centralized training with decentralized execution (CTDE) and independent learning (IL) methods \citep{zhang2020taxonomy, Gronauer2021, anniemultiagent}. These algorithms belong to two categories: Q-learning-based algorithms, such as Independent Q-Learning (IQL) \citep{IQL}, Value-Decomposition Networks (VDN) \citep{VDN}, and QMIX \citep{QMIX}, and PG algorithms, such as Multi-Agent Proximal Policy Optimization (MAPPO) \citep{yu2022surprising} and Multi-Agent Advantage Actor-Critic (MAA2C) \citep{foerster2018counterfactual}. 
We employed \textit{parameter sharing} across all algorithms during training which was shown by \cite{papoudakis2021benchmarking} to lead to greater episode returns when compared to cases without parameter sharing.



\section{Case Studies}
\begin{figure}[ht!]
    \centering
    \begin{subfigure}[b]{0.23\textwidth}
        \includegraphics[width=\textwidth]{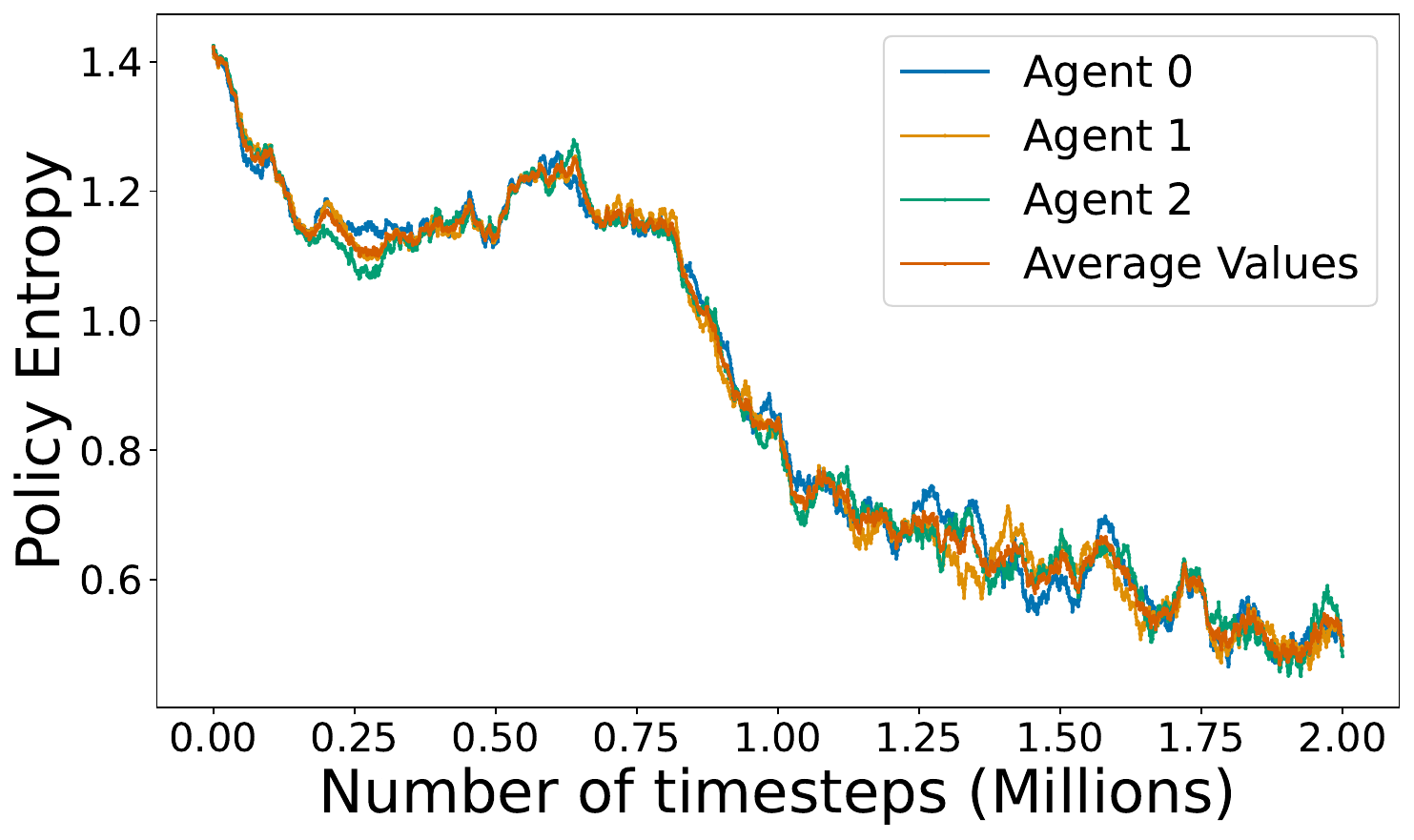}
        \caption{MAA2C}
        \label{fig:maa2c_entropy}
    \end{subfigure}
     \begin{subfigure}[b]{0.23\textwidth}
        \includegraphics[width=\textwidth]{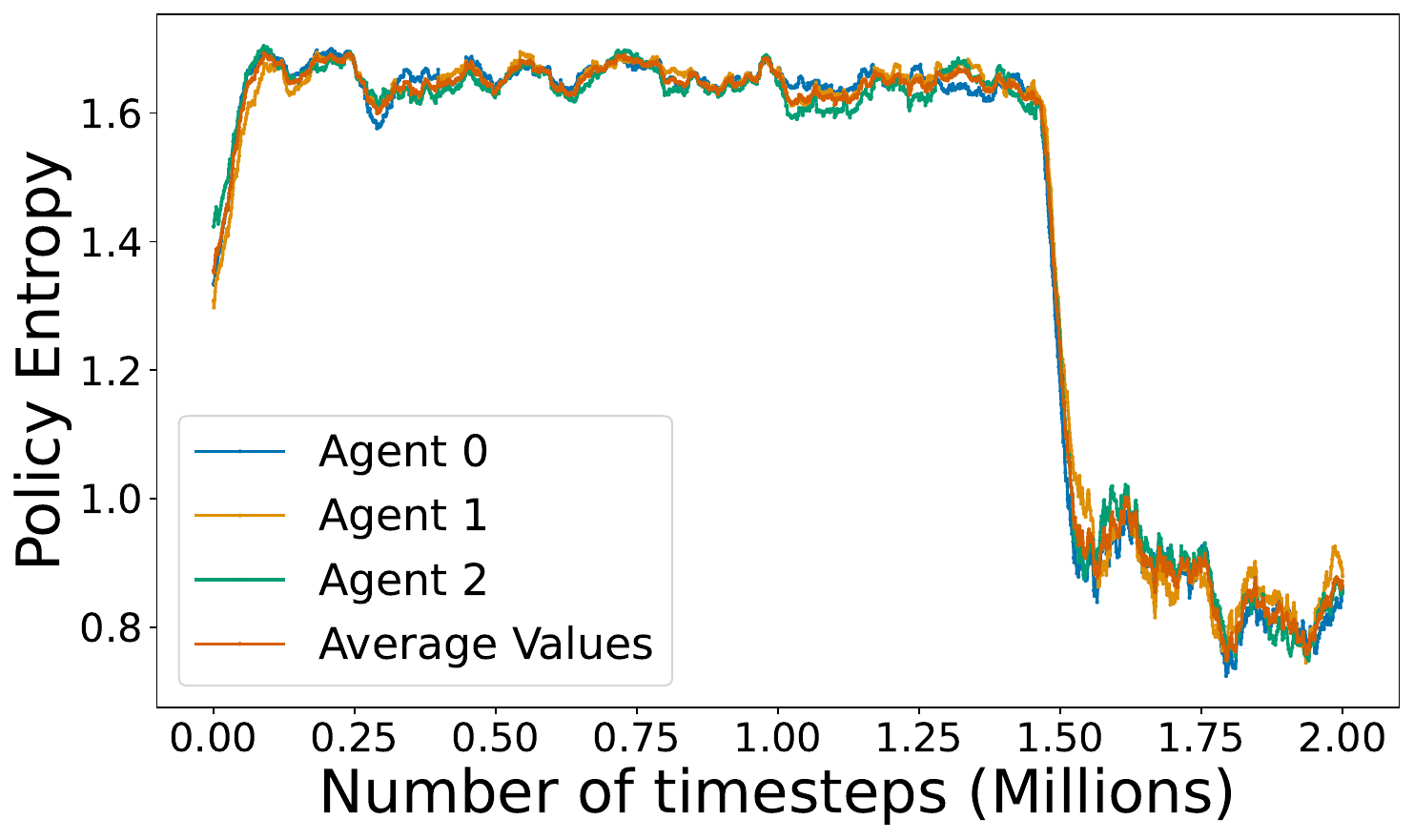}
        \caption{MAPPO}
        \label{fig:mappo_entropy}
    \end{subfigure}
    \hfill
    \begin{subfigure}[b]{0.23\textwidth}
        \includegraphics[width=\textwidth]{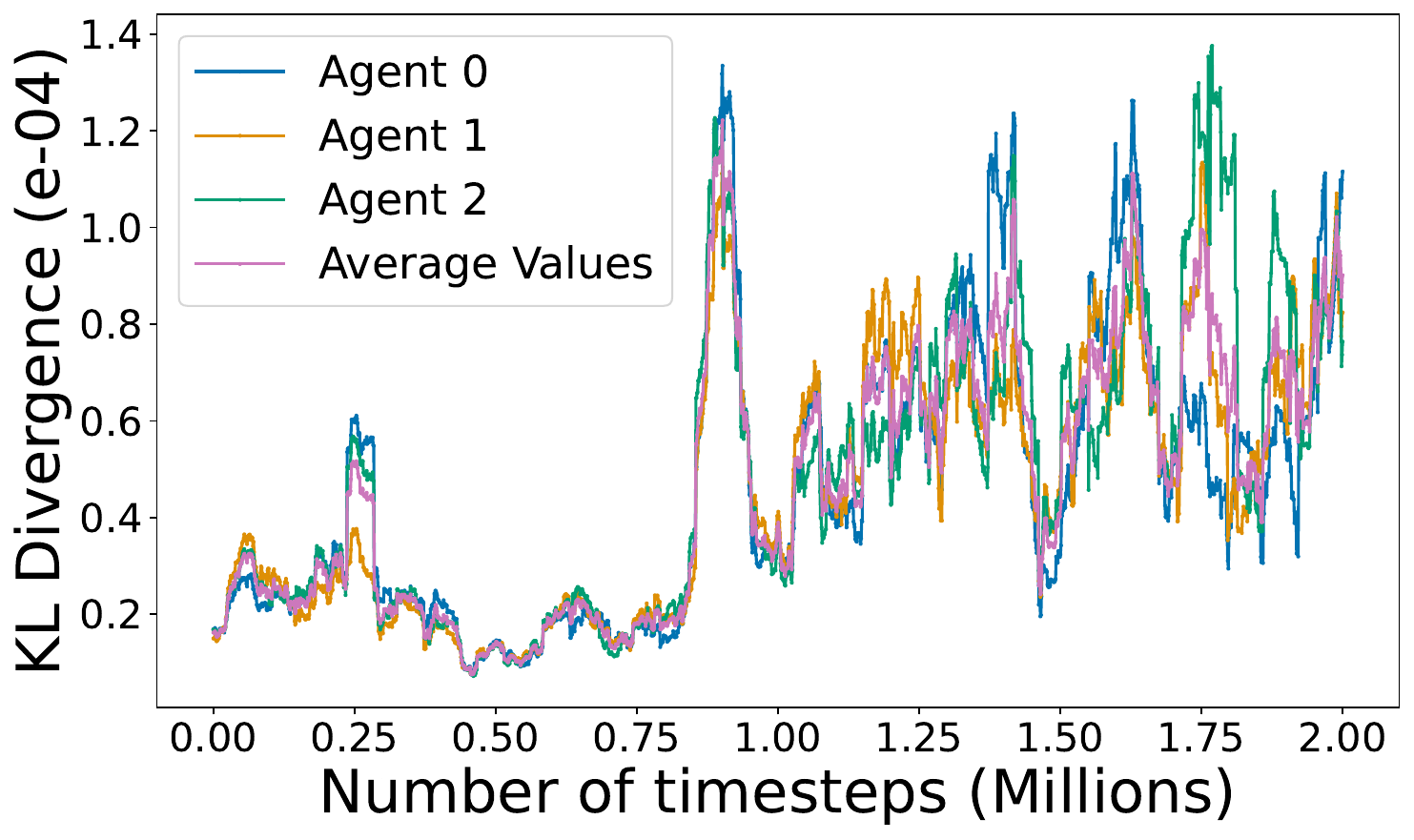}
        \caption{MAA2C}
        \label{fig:maa2c_kl_div}
    \end{subfigure}
    \begin{subfigure}[b]{0.23\textwidth}
        \includegraphics[width=\textwidth]{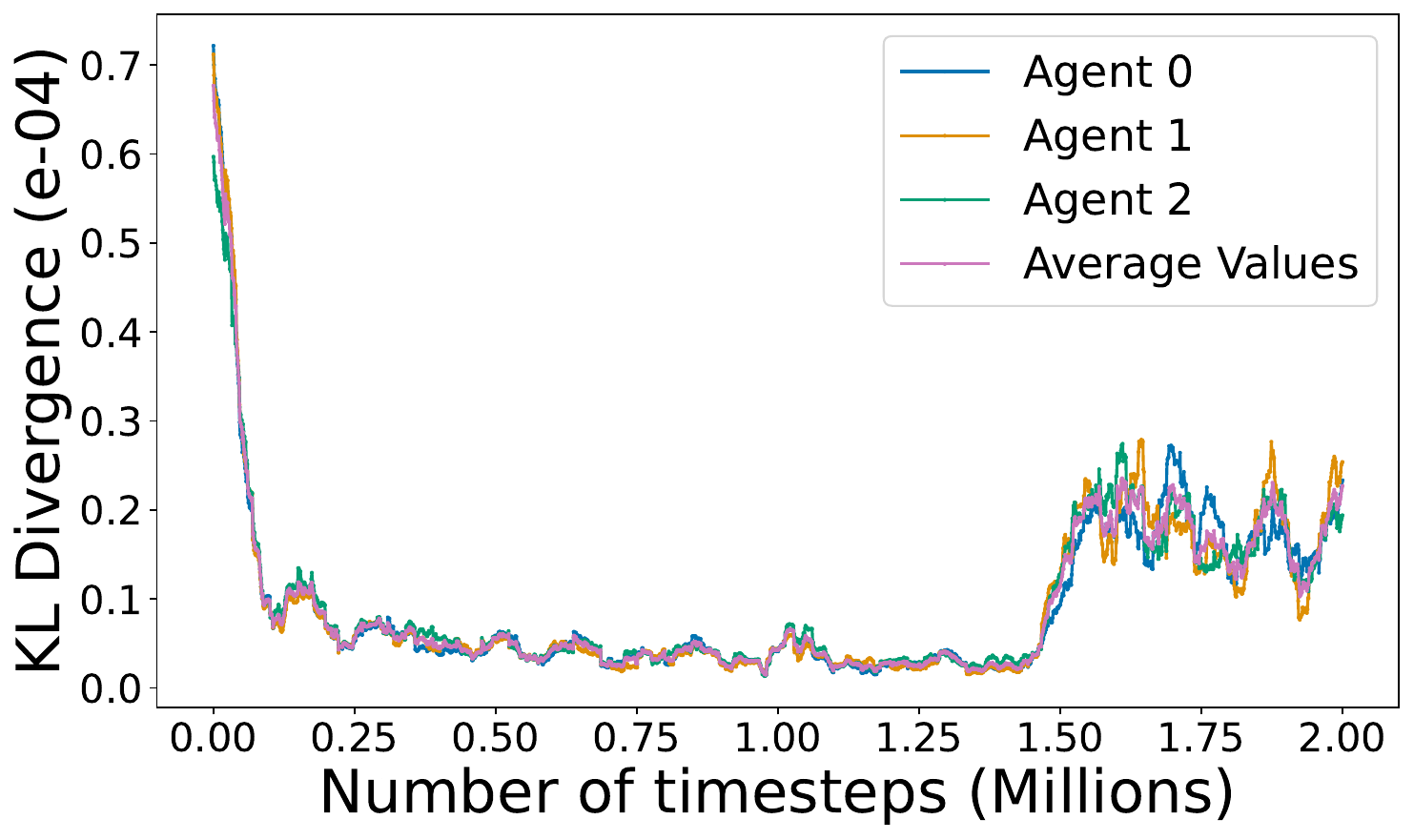}
        \caption{MAPPO}
        \label{fig:mappo_kl_div}
    \end{subfigure}
    \caption{\textit{Training stability results on \textit{Foraging-15x15-3p-5f} (one seed)}. \textbf{Top row:} Policy Entropy. \textbf{Bottom row:} Agent Update Divergence.}
    \label{fig:kl_entropy_mappo_maa2c}
\end{figure}

\begin{figure}[ht!]
    \centering
    \begin{subfigure}{0.45\textwidth}
        \includegraphics[width=\textwidth]{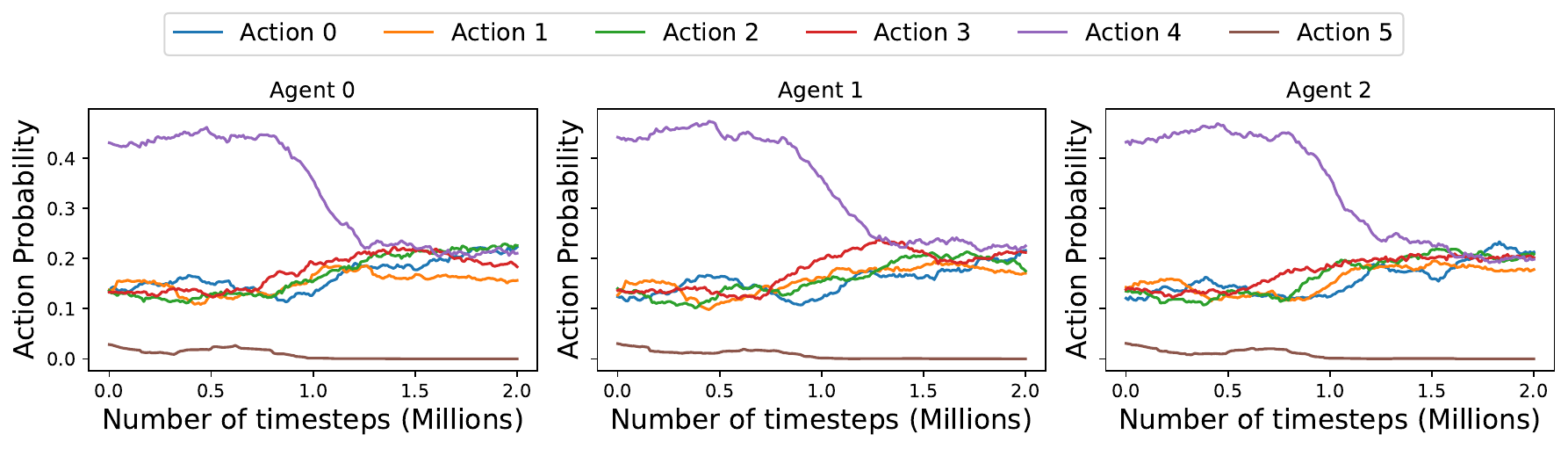}
    \end{subfigure}
    \hfill
    \begin{subfigure}{0.45\textwidth}
        \includegraphics[width=\textwidth]{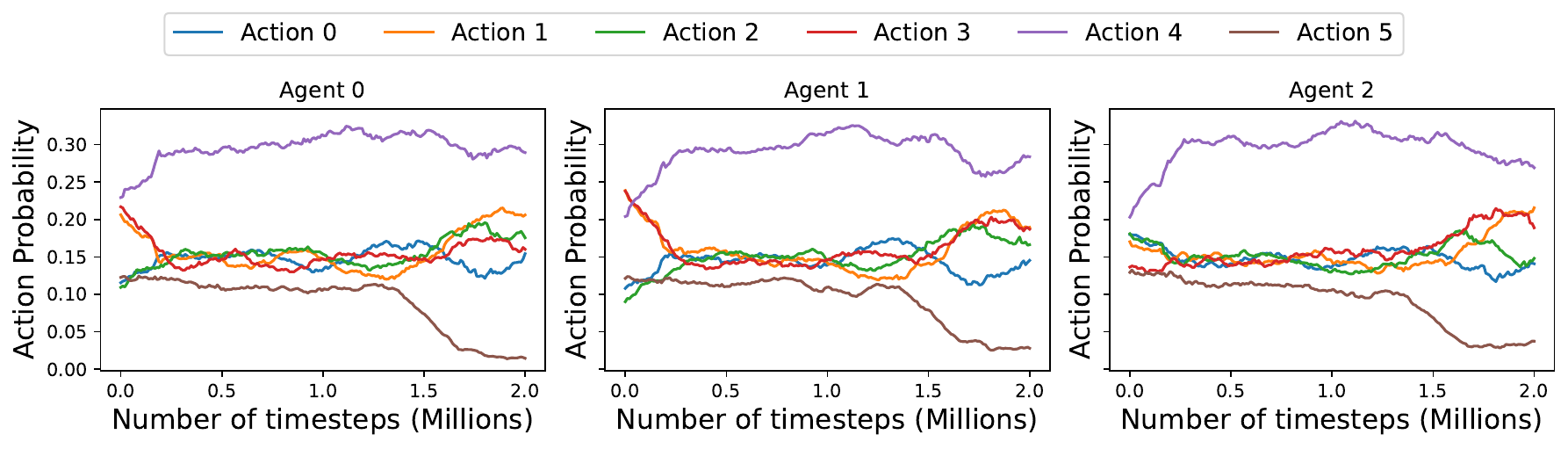}
    \end{subfigure}
    \caption{\textit{Task Switching on \textit{Foraging-15x15-3p-5f} (one seed)}. \textbf{Top:} MAA2C. \textbf{Bottom:} MAPPO.}
    \label{fig:ts_mappo_maa2c}
\end{figure}

\begin{figure}[ht!]
     \centering
     \begin{subfigure}[t]{0.5\textwidth}
         \includegraphics[width=\textwidth]{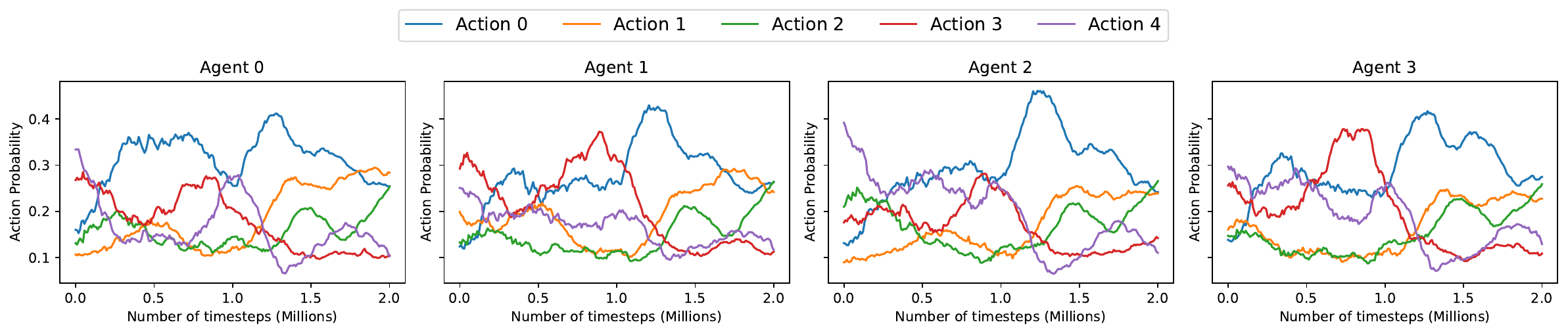}
         \captionlistentry{}
         \label{fig:rware_task}
     \end{subfigure}
     \hfill
     \begin{subfigure}[t]{0.5\textwidth}
         \includegraphics[width=\textwidth]{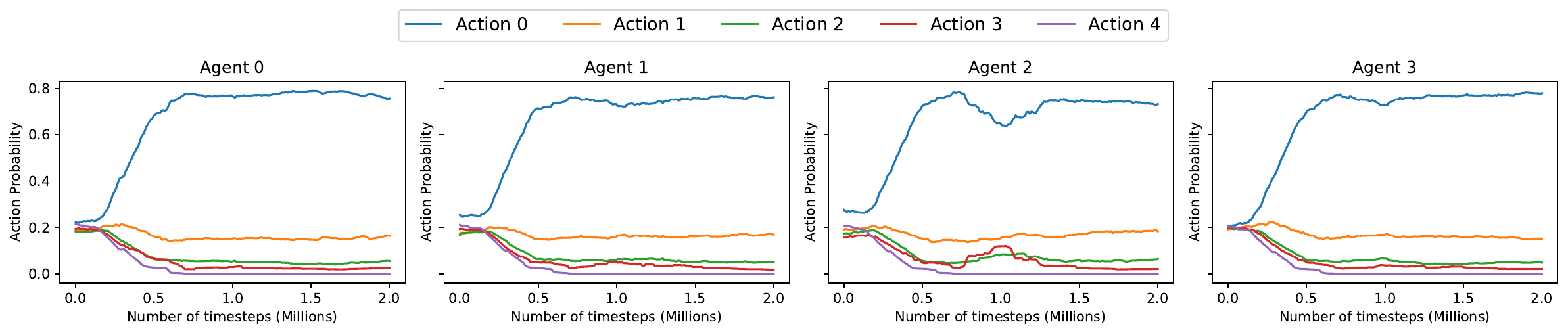}
         \captionlistentry{}
         \label{fig:lbf_task}
     \end{subfigure}
     \caption{\textit{Comparison of task-switching for MAPPO and IQL algorithms on the RWARE setting} \textbf{Top:} IQL. \textbf{Bottom} MAPPO.}
     \label{fig:iql_mappo_rware}
 \end{figure}

After running experiments on seven scenarios in the LBF environment, each with ten different seeds, our results point out that relying solely on empirical returns may not offer a complete understanding of algorithmic performance in MARL. However, when complemented by simple XAI methods, these results can provide valuable insight into the environmental and algorithmic factors of the MARL setting. 
In the following, from Figure \ref{fig:kl_entropy_mappo_maa2c} to \ref{fig:iql_mappo_rware}, different agents have similar behaviour which indicates that they have learned similar or redundant policies that do not vary much across scenarios and this outcome can be attributed to the use of parameter sharing in the learning process. 

\textbf{Why does MAA2C outperform MAPPO? }We attempt to answer this question using the above diagnostic. We focus on the LBF scenario \textit{Foraging-15x15-3p-5f}, using one seed, to examine the performance disparities between MAA2C and MAPPO. Based on the fact that the core architecture of PPO \citep{yu2022surprising} extends A2C by introducing a clipping function, it is plausible to attribute the reduced performance of MAPPO to its sensitivity to the clipping ratio, which can lead to an early suboptimal policy. A smaller clipping ratio can make PPO overly cautious, slowing down the learning process. Conversely, a larger clipping ratio can make PPO overly aggressive, potentially destabilising the learning process. 
This hypothesis is supported by the slower convergence of the MAPPO algorithm across all LBF scenarios, which initially struggles to improve its policy, as shown in Figure \ref{fig:kl_entropy_mappo_maa2c} and \ref{fig:ts_mappo_maa2c}. Furthermore, MAPPO continues to face challenges in reaching a performance level comparable to MAA2C in the LBF environment, as shown in Figure \ref{fig:benchmarking_paper_redo}, where the probability that MAA2C outperforms MAPPO in all LBF scenarios on 10 different seeds is close to 1.

The learning dynamics, characterised by the Agent Update Divergence and Policy Entropy metrics, exhibit notable differences between agents trained under the MAA2C and MAPPO algorithms. Specifically, MAA2C demonstrates a more effective learning trajectory, marked by consistent steps in policy improvement.\\
In Figures \ref{fig:maa2c_entropy} and \ref{fig:maa2c_kl_div}, as well as in the average values across all agents, we observe distinct patterns for agents trained with MAA2C. Initially, their learning progresses slowly, but subsequently, their policies undergo significant changes. This phenomenon can be attributed to continuous exploration of the environment, leading to the discovery of more effective strategies. Moreover, the increased certainty in action selection suggests successful policy updates. These metrics collectively indicate that the policies remain stochastic and exploratory, evolving frequently over time without full convergence. \\
In contrast, when examining the behaviour of agents trained with the MAPPO algorithm in Figures \ref{fig:mappo_entropy} and \ref{fig:mappo_kl_div}, we observe a different pattern. It appears that the sensitivity to the clipping ratio in MAPPO may hinder the agents from converging rapidly, in addition to the fact that the environment is highly random (levels and coordinates). The policy may become trapped in a local optimum early in training, persisting for over one million steps. This situation limits the algorithm's ability to explore effectively and discover a superior policy during the initial stages of training.\\
In Figure \ref{fig:ts_mappo_maa2c}, action 0 corresponds to the "no operation" action, action 5 is used to pick up nearby food, and the remaining actions are movement actions.
The \textit{Task Switching} plots reveal that MAA2C becomes more sensitive to changes in action selection, as indicated by the converging action probabilities. This enhanced sensitivity allows MAA2C to explore and utilise a broader range of actions, leading to a reduction in actions that are associated with lower rewards. In contrast, MAPPO exhibits greater action dominance, particularly in one action (action 4). The differences in performance between MAA2C and MAPPO can be attributed to their distinct learning strategies.


\textbf{Sparsity or something else?} From figure \ref{fig:benchmarking_paper_redo} it would seem that Q-learning methods are almost completely unable to learn in the RWARE setting. \cite{papoudakis2021benchmarking} reasonably hypothesise that this is due to the sparsity of the setting making value decomposition difficult for VDN and QMIX and that IQL is hampered by its limited exploration capabilities. However, both VDN and QMIX have shown a limited capacity to learn some sparse settings \citep{mguni2022ligs} and IQL is in some cases able to find policies that can solve highly complex sparse settings in the MARL domain \citep{mahajan2019maven}. From figure \ref{fig:iql_mappo_rware}, we can find an alternative explanation for the poor performance of the Q-learning methods. For the high-performing policies found by MAA2C and MAPPO, we can see that there is a high reliance on action 0 which has been revealed by the Task Switching measurement. This indicates that the performance issues are potentially not purely due to sparsity but the Q-learning methods not learning to exploit a specific action. As Q-learning methods use a replay buffer to train from off-policy data it is possible that they do not encounter samples exploiting this action in combination with the relevant states frequently enough which makes learning the heavily weighted optimal policy difficult. Manually parsing environment trajectories to find this data would be difficult and this finding shows how even simple XAI methods can make algorithmic failure points and environment features easier to detect.

\section*{Conclusion}

This paper highlights the importance of explainability in MARL research. While traditional performance metrics may not fully reveal the true behaviours of agents, we demonstrate how diagnostic tools—specifically, Policy Entropy, Agent Update Divergence, and Task Switching— can provide a more comprehensive overview of agents' behaviour in multi-agent systems. 
Our work contributes to the growing field of explainable MARL by providing a set of simple yet effective tools that can be applied to various MARL algorithms and environments. Further exploration of explainability in MARL is encouraged to address challenges and fully harness its potential in solving complex problems.
It is essential to acknowledge the \textit{limitations} of our work, including the computational overhead that escalates with an increasing number of agents and the potential inefficiency of the Task Switching tool in scenarios with continuous action spaces.
Looking ahead, we envision \textit{future research} delving deeper into the application of these tools in more  MARL environments. Expanding and refining these diagnostic tools will benefit both researchers and practitioners in the field, pushing the boundaries of explainable multi-agent reinforcement learning and enhancing our ability to tackle complex real-world challenges.

\bibliography{references}
\end{document}